\documentclass{article}

    \usepackage[preprint]{neurips_2024}

\usepackage[utf8]{inputenc} 
\usepackage[T1]{fontenc}    
\usepackage{hyperref}       
\usepackage{url}            
\usepackage{booktabs}       
\usepackage{amsfonts}       
\usepackage{nicefrac}       
\usepackage{microtype}      
\usepackage{xcolor}         
\usepackage{multirow}
\usepackage{multicol}
\usepackage[normalem]{ulem}
\useunder{\uline}{\ul}{}
\usepackage{wrapfig}
\usepackage{svg}
\usepackage{caption}
\usepackage{flushend}
\usepackage{amsmath}
\usepackage{amssymb}
\usepackage{colortbl}
\usepackage[switch]{lineno}
\usepackage{adjustbox}

\title{Evaluating Hallucination in Text-to-Image Diffusion Models with Scene-Graph based Question-Answering Agent}

\author{
Ziyuan Qin$^{1, 2}$\thanks{Equal contribution}\quad 
Dongjie Cheng$^{1*}$\quad 
Haoyu Wang$^{5}$\quad 
Huahui Yi$^{1}$\AND
Yuting Shao$^{2}$\quad 
Zhiyuan Fan$^{6}$\quad
Kang Li$^{1,3}$\quad
Qicheng Lao$^{3,4}$\thanks{Corresponding author}\\ \\
$^1$West China Biomedical Big Data Center, West China Hospital, Sichuan University\\
$^2$Case Western Reserve University, OH\\
$^3$School of Artificial Intelligence, BUPT, Beijing, China\\
$^4$Shanghai Artificial Intelligence Laboratory, Shanghai, China \\
$^5$Rockontrol Lab, Chengdu, China \\
$^6$Tianjin University, Tianjin, China \\
\texttt{ziyuan.qin@case.edu}
}

\begin{document}

\maketitle

\begin{abstract}
Contemporary Text-to-Image (T2I) models frequently depend on qualitative human evaluations to assess the consistency between synthesized images and the text prompts. There is a demand for quantitative and automatic evaluation tools, given that human evaluation lacks reproducibility. We believe that an effective T2I evaluation metric should accomplish the following: detect instances where the generated images do not align with the textual prompts, a discrepancy we define as the `hallucination problem' in T2I tasks; record the types and frequency of hallucination issues, aiding users in understanding the causes of errors; and provide a comprehensive and intuitive scoring that close to human standard.
To achieve these objectives, we propose a method based on large language models (LLMs) for conducting question-answering with an extracted scene-graph and created a dataset with human-rated scores for generated images. From the methodology perspective, we combine knowledge-enhanced question-answering tasks with image evaluation tasks, making the evaluation metrics more controllable and easier to interpret. For the contribution on the dataset side, we generated 12,000 synthesized images based on 1,000 composited prompts using three advanced T2I models. Subsequently, we conduct human scoring on all synthesized images and prompt pairs to validate the accuracy and effectiveness of our method as an evaluation metric. All generated images and the human-labeled scores will be made publicly available in the future to facilitate ongoing research on this crucial issue. Extensive experiments show that our method aligns more closely with human scoring patterns than other evaluation metrics.
\end{abstract}

\section{Introduction}
As John Powell insightfully remarks, ``The only real mistake is the one from which we learn nothing.'' Recognizing and critically analyzing our errors lays the foundation for substantial progress. However, without establishing a reasonable evaluation metric, how can we learn from our mistakes? For a long time, the field of text-to-image (T2I) generation has faced the challenge of lacking cost-effective and efficient evaluation methods. Researchers dedicated to improving T2I models have had to rely on expensive and limited human subjective scoring to glean only a small amount of insight to enlighten them on the path of improvement.

Early image evaluation metrics, such as SSIM and PSNR~\cite{Hor2010ImageQM}, focused more on the visual quality of synthesized images and were unable to measure the consistency between text prompts and images. With the advent of vision-language models (VLMs) like CLIP~\cite{Radford2021LearningTV} and BLIP~\cite{Li2022BLIPBL}, comparing the similarity of embedding vectors between generated images and text prompts has become a mainstream evaluation metric. However, the limitation of this method is that it provides a coarse-grained, whole-image level similarity calculation, which is insensitive to certain detailed errors. Furthermore, similarity-based metrics lack interpretability and offer just an abstract number rather than concrete guidance.
The inadequacy of coarse-grained evaluation metrics becomes particularly evident when it meets complex composite prompts. Composite textual prompts often contain multiple objects, each potentially having more than one modifying attribute. In such scenarios, T2I models are prone to issues of attribute mismatch.

A few recent works aim to tackle this challenging problem in T2I evaluation~\cite{Huang2023T2ICompBenchAC,Hu2023TIFAAA,Lee2023HolisticEO}. They use modularized evaluation framework to run several independent tests regarding various aspects of the synthesized images. In T2ICompBench~\cite{Huang2023T2ICompBenchAC} proposed to use the probability of answering ``yes'' for a question about the attribute or relations of objects as the score for the attribute binding evaluation. Holistic Evaluation of T2I Models (HEIM) splits its evaluation into 12 independent departments. However, it cannot provide a comprehensive score for each image. Thus, we need more advanced evaluation metrics to effectively capture the hallucination issues in T2I tasks. Hallucination is a term that has frequently appeared along with AI-Generated Content (AIGC) recently. It refers to the phenomenon that generative models create content inconsistent with the prompts or contain factual errors. In the T2I context, we define hallucination as the inconsistency existing between synthesized images and text prompts. 
As highlighted in our abstract, we assert that comprehensive evaluation metrics for T2I tasks should encompass the following aspects: 1. Identify hallucinations in synthesized images at various levels, including object-level, attribute-level, and relationship-level, etc. 2. Classify the types of hallucinations and quantify their occurrences. 3. Provide an overall assessment of the consistency between the generated images and the textual prompts.

For the purpose of innovating a more comprehensive evaluation metrics, we first build a dataset with a large volume of images generated by T2I models from a list of complicated composite text prompts curated by ~\citep{Huang2023T2ICompBenchAC}. We then hand these text prompts to human evaluators and provide them with guidelines on how to evaluate these generated images. These human-generated scores and the synthesized images will be made public so other researchers can use this dataset to evaluate their own metrics.

Additionally, we also present an innovative algorithm capable of integrating multiple evaluation dimensions into a unified framework, thereby furnishing a comprehensive quantifiable metric. Our approach integrates open object detection methods and VQA models to extract a scene-graph from images. We then generate template questions and answer pairs from the dependency tree of the text prompts, focusing on the existence of objects, their relationship, and whether they possess certain attributes. Finally, we use the aforementioned knowledge graph as context and employ an LLM model to answer our posed questions. We hypothesize that a higher accuracy in responses based on our extracted scene graph indicates a greater alignment of the generated image with the original textual prompts. 


From out experiments, except for explicitly unmatched relationships, objects, or attributes, we found that the generative models tend to generate some unmentioned content to enrich the images. Specifically, if the synthesis models are instructed to generate a table, they will not generate an image with just one table. To enrich the frame, the models will arbitrarily include more details, such as chairs and cutlery. 
Our algorithm can detect these `extraneous objects. We attempted to analyze which unmentioned objects are necessary or acceptable and which are disruptive. We also employed manual scoring to identify unnecessary elements. However, due to the subtlety of these extraneous objects, even human raters have not reached a consensus, making it challenging to provide an automated evaluation of this type of hallucination. Thus, we can only conduct a qualitative analysis of this issue and expect future works to tackle this problem. 

Overall, the key contributions of our work are fourfold: 
\begin{itemize}
  \item We propose a method to evaluate T2I generative models through question-answering based on a local knowledge graph constructed from information extracted from the images. This approach effectively assesses whether the generative models faithfully reflect the provided textual prompts and identifies the types of hallucinations produced in the generated images.
  \item  As a groundbreaking study concentrating on the hallucination phenomenon in T2I generative models, we identify and define the hallucination phenomenon in contemporary text-generated image content. Furthermore, we categorize these hallucinations and explore their potential causes. This feature brings more interpretability to our method.
    \item  We establish criteria based on human evaluation results to determine whether these `extraneous objects' constitute an unreasonable hallucination. Extensive experiments show that our rating align with the human scoring.
  \item We perform a gradation of the illusion phenomena in our text-to-image content and establish a scoring metric. Using three different text-to-image generative models, we generated a total of 12,000 images from 1,000 prompts, with annotators comprehensively scoring these images according to the scoring rules we made. Both generated images and human-labeled scores will be publicly available for the community to facilitate further research.

\end{itemize}

\section{Related Work}
\label{relatedwork}
\subsection{The development of Text-to-Image Models}
The task of T2I generation has always been a challenge in the field of image synthesis, as it involves understanding text and reflecting this understanding in synthesized images. Early works incorporated textual vectors as conditions into Variational Autoencoder (VAE) models~\cite{yan2016attribute2image}. However, this approach required pre-encoding a closed set of attribute texts and did not support free form text-to-image tasks. VQVAE~\cite{vqvae,vqvae2,ramesh2021zeroshot}, an advancement over the traditional VAE framework, incorporates vector quantization to achieve a discrete latent representation. The author posits that this discrete latent representation is more apt for Text-to-Image (T2I) tasks, given that text inherently exists in a discrete format. Recently, the rise of diffusion models has marked a significant leap forward in the field of image synthesis. Diffusion models~\cite{Rombach2021HighResolutionIS,Saharia2022PhotorealisticTD,Ramesh2022HierarchicalTI,Ho2020DenoisingDP} initially add noise into an image through forward propagation, and during training, they reverse this process to restore the original image from its noised state. Compared to previous models, diffusion models require multiple iterative steps in the synthesis process to produce the final image. Subsequent works like the Latent Diffusion Model (LDM)~\cite{Saharia2022PhotorealisticTD,Rombach2021HighResolutionIS} have integrated a conditioner into the model architecture. This conditioner can accept various prompts, including text, sketches, or bounding boxes. With the introduction of the conditioner, diffusion models have gained the capability to perform T2I tasks with any form of text, achieving results comparable to human-produced images. Since then, numerous publicly accessible diffusion-based models have emerged within the AI community. The Stable Diffusion Model (SDM) series is among the most renowned diffusion-based model series currently available on the market. In this study, we utilize three different versions of SDMs to generate images from text prompts. The SDv2 (Stable Diffusion version 2) model replaces the original OpenAI's CLIP~\cite{Radford2021LearningTV} text encoder in SDv1-4(Stable Diffusion version1.1-4) with OpenCLIP~\cite{ilharco_gabriel_2021_5143773} as its text encoder. SDxl~\cite{podell2023sdxl} further incorporates a refiner module and adds a second text encoder to achieve enhanced generative performance.

\subsection{Hallucination in LLM and VLM}
Hallucination issues refer to content generated by models that contradicts the given instructions, fails to meet the requirements, or deviates from factual accuracy. Such hallucination issues have been identified and studied in LLMs and VLMs~\cite{zhang2023sirens,Zhou2023AnalyzingAM,Yin2023WoodpeckerHC}, but have not been thoroughly explored in T2I generation models. Previous studies~\cite{zhang2023sirens,Yin2023WoodpeckerHC} on hallucination phenomena have mainly focused on discrepancies between content and prompts, factual inaccuracies, and inconsistencies in content. Due to the specific nature of T2I tasks, our current study will not delve into the latter two issues, focusing primarily on the problem of discrepancies between content and prompts. 

\subsection{Other Evaluation Metrics for Image Synthesis}
Early image evaluation metrics focused on assessing the quality of generated images. SSIM (Structural Similarity Index Measure) and PSNR (Peak Signal-to-Noise Ratio)~\cite{Hor2010ImageQM} accomplish this task by directly comparing synthesized images to original images at the pixel level. Inception Score (IS) and FID (Fréchet Inception Distance) Score~\cite{Chong2019EffectivelyUF,Barratt2018ANO} involve the use of a classifier pre-trained on a large dataset to compare the differences between synthesized and original images. The R-precision~\cite{Kynknniemi2019ImprovedPA} metric transforms the task of evaluating synthesized images into the task of recalling the original text prompts with the embedding of the images. R-CLIP-Precision~\cite{Park2021BenchmarkFC} further refines this by utilizing the CLIP model to encode images, thereby achieving more accurate recall scores. CLIP-score~\cite{Hessel2021CLIPScoreAR} also adopts the pre-trained CLIP model to obtain the embedding of the synthesized images and text prompts. The other line of work~\cite{Lu2023LLMScoreUT,Lee2023HolisticEO} leverages pre-trained VLMs to generate captions of the synthesized images and compare the similarities between the captions and the original prompts. VQA-based methods~\cite{Hu2023TIFAAA,Huang2023T2ICompBenchAC,DBLP:journals/corr/abs-2202-04053} directly query the VQA models to obtain the attributes, relations, and other aspects of the synthesized images and evaluate the images according to both the answers and the text prompts. Our method moves a further step, building a local knowledge graph from the synthesized images and employ an LLM model to do graph-based question-answering tasks for evaluation.

\section{Method}
\label{method}
\subsection{Identify the unmentioned objects}
\begin{figure}[t]
    \centering

    \includegraphics[width=1.0\textwidth]
    {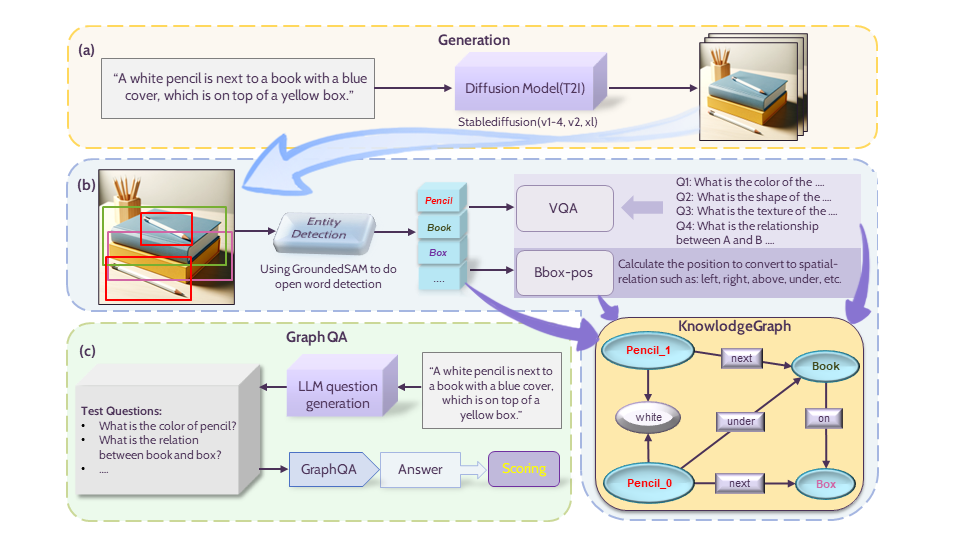}
    
    \caption{The complete framework for the text-to-image (T2I) generation to evaluation process is as follows: (a) First, we generate images from the textual prompts; (b) Then, we extract entities from these images and construct a knowledge graph; (c) Subsequently, we generate template questions based on the textual prompts and perform GraphQA using the constructed knowledge graph. Finally, we score the answers to obtain the final evaluation score.}
    \label{fig:fig1}
\end{figure}

The whole pipeline of the evaluation framework is depicted in Figure~\ref{fig:fig1}. We first employ pre-trained T2I generative models to generate images based on text prompts. Then, we use detection models and VQA models to build an attribute-aware knowledge graph that reflects all crucial information in the generated image. After we build the local knowledge graph from the image, we create a set of template questions from the text prompts, covering all objects, relations, and attributes in the prompt. After all, we utilize an LLM model to answer the question we raised before by exploiting the knowledge graph. The model will examine the knowledge graph to make answers. And finally, the rule-based scoring module will calculate the final scores according to the answers made before. The entire process requires no human intervention or any training. 

\subsection{Building Image-specific Scene-Graph}
Building an image-specific scene-graph from an image is a non-trivial task, and many existing works leverage graph neural networks(GNN) to do it. However, considering the cross-modal nature of building the knowledge graph, we decided to leverage large pre-trained VLMs to extract fine-grained semantic information from the image accurately. Specifically, we first use an open-vocabulary detection algorithm GroundedSAM~\cite{Grounded-SAM_Contributors_Grounded-Segment-Anything_2023} to detect all salient objects in the images. Subsequently, we cropped each detected region and fed them into BLIP-2~\cite{Li2023BLIP2BL} to perform VQA tasks. During the VQA process, we asked the model about the cropped area's color, shape, and texture attributes. For each detected object, there is a corresponding node in the knowledge graph to represent it. Attributes will considered as nodes that bind to the objects they modify. Through the aforementioned approach, we successfully constructed all the nodes in the graph, and the next step is to identify the edges connecting these nodes. 
As shown in Figure~\ref{fig:fig1}(a), for non-spatial relations, we just feed the BLIP-2 model with two cropped regions along with their corresponding object names and then ask the model to return the relationship between them. As for spatial relations, we follow the routine of using the bounding box coordinates to determine the spatial relationship between objects.
Another issue in building the graph from an image is the redundancy of edges. We should not expect the scene graph to be fully connected, so we filter out some unnecessary edges by performing similarity checks on the names of two nodes.

\subsection{Rasing Templated Questions via Dependency Tree}
To evaluate whether a generated image aligned with the text prompt, we want to know whether the images contain the presence, attributes, and relationships among objects mentioned in the prompts. To achieve that, we first want to raise some questions from the text prompts about the attributes, relationships, and objects.
We use Spacy~\cite{Honnibal_spaCy_Industrial-strength_Natural_2020} package to do Dependency Parsing on the text prompts and only keep the relevant components from the parsed segments. Formally, dependency parsing will break a prompt into tokens and return each token's grammatical dependency and part-of-speech (POS) tag. After parsing, every word is a node in the dependency tree, and each grammatical dependency is an edge, having a depend head and a type. We can split words into small groups by the grammatical dependency, and the verbs or prepositions will bridge these groups. Once we obtain the tree, we employ LLMs to generate questions and corresponding answers regarding the relation and attributes of objects. We formulate the nodes and edges in the dependency trees into triples. For example: `Subject: Cat -- Prepositions: on the top of -- Object: Sofa' represents the spatial relation between cat and sofa in the dependency tree. We feed these triples as context to the LLMs and ask them to generate questions and answers regarding the relations and attributes according to these triples. For instance, the LLM will raise a question about the relationship between the cat and the sofa in the triples we presented before, and the answer to that question is the relation in the triple. By providing several in-context demonstrations, LLM can easily generate near-perfect question-answer pairs based on contextual triples. The specific demonstration templates can be found in our supplementary materials.

\subsection{Evaluation as Question-Answering Task}
We formulate the evaluation problem into a scene-graph-based question-answering task on the knowledge graph we built before, while the questions are from the process we mentioned in the last section. As depicted in Figure~\ref{fig:fig1}(c), on the left side is the local knowledge graph we constructed previously. The questions we formulated based on the text prompts are on the right side. To evaluate whether the generated images reflect the requirement in the text prompts, we employ an LLM model to answer these questions raised from the text prompts based on the scene-graph extracted from the synthesized images. The LLM model retrieved relevant triples from the scene-graph to verify the details, such as attributes, relationships, etc., and make answers. Concretely, we iteratively ask the LLM model to answer the questions we generated in the last section while providing the whole knowledge graph as context information. We use LangChain GraphQA module~\cite{Chase_LangChain_2022} to perform this task. 
A GraphQA agent initially extracts entities from the question and retrieves triples associated with these entities from the Knowledge Graph. These triples constitute the 'Memory' for this agent. Subsequently, the agent assesses whether the retrieved triples sufficiently address the posed question.

\begin{gather}
    \label{eq:eq1}
    mem = R(ent|KG;Agent), ent = Ext(q) \\
    Ans = Decoder(mem, q|Agent) 
\end{gather}

In the above equations, $R$ represents the retrieval model, while $Ext$ refers to the entity extraction module. The extracted $ent$ signifies the entities in question $q$. Ultimately, the recalled $Mem$ denotes the 'Memory' of the Agent, as mentioned earlier. By inputting both the Memory and the question $q$ into the decoder, the agent will return an answer $Ans$.


\begin{table*}[ht]
\tiny
\caption{Comparing different Evaluation Benchmark with Human Scores: The left half of the table displays the scores given by different evaluation metrics to the three generative models. A darker background color indicates a higher score, and vice versa. Apart from our method, we can observe that the scoring trends demonstrated by other methods do not align with human scoring pattern.} 
\label{table:table1}
\vspace{-10pt}
\begin{center}
\resizebox{1\linewidth}{!}{%
\begin{tabular}{cccc|ccc}
\toprule
\textbf{Metrics} & $Stable_{v1-4}$ & $Stable_{v2}$ & $Stable_{xl}$ & Pearson & Kendall's $\tau$ & Spearman's $\rho$\\
\midrule
CLIP-Score  & \cellcolor{red!50}0.62 & \cellcolor{red!15}0.48  & \cellcolor{red!25}0.54 & 0.14 &0.10 & 0.13 \\
BLIP-Score & \cellcolor{red!15}0.6461 & \cellcolor{red!50}0.75 & \cellcolor{red!25}0.72 & 0.12 & 0.06 & 0.09 \\
LLM-Score  & \cellcolor{red!15}0.63 & \cellcolor{red!50}0.65 & \cellcolor{red!25}0.64 & 0.19 & 0.13 & 0.18\\
T2ICompbench & \cellcolor{red!15}0.29 & \cellcolor{red!50}0.30 & \cellcolor{red!15}0.29 & 0.24 & 0.20 & \textbf{0.27}\\
\midrule
Ours(w/Gemini) & \cellcolor{red!25}0.36 & \cellcolor{red!15}0.28 & \cellcolor{red!50}0.41 & 0.23 & 0.21 & 0.25\\
Ours(w/GPT-4) & \cellcolor{red!25}0.47 & \cellcolor{red!15}0.46 & \cellcolor{red!50}0.49 & \textbf{0.26} & \textbf{0.23} & \textbf{0.27}\\
\midrule
Human Score & \cellcolor{red!25}0.64 & \cellcolor{red!15}0.63 & \cellcolor{red!50}0.69 & - & - & -\\
\bottomrule
\end{tabular}
}
\end{center}
\end{table*}

\subsection{Records Hallucination Problem Types}
In conclusion, the core assumption is that if the generated images faithfully reflect the content in the text prompts, the graph question-answering agent should be able to answer all the questions by retiring the relevant triples in the scene-graph. Otherwise, an erroneous answer may suggest a hallucinated T2I content.

During the GraphQA process, we record every incorrect answer made by the GraphQA agent and categorize this error into three divisions: 1. Attribute Hallucination; 2.Relation Hallucination; 3. Object Hallucination. The first kind is attribute-level hallucination, which means some objects have incorrect and unexpected attributes. The second kind is relationship-level hallucination, which always refers to the case that we answer incorrect relationships between objects. The last kind is called object-level hallucination, which includes both the existence of the unmentioned objects and the absence of the mentioned objects in the text prompts. After the GraphQA module answers all the questions, we feed the record of incorrect answers into a rule-based scoring system to generate a final rating for the image. 

\section{Experiments and Analysis}
\subsection{Dataset Creation Detail}

For the experiments, we use 3 different T2I models to generate 12,000 images in total with respect to 1000 novel composite text prompts. These prompts are from T2I-CompBench~\cite{Huang2023T2ICompBenchAC} released prompts for complex composition scenarios. In a nutshell, each one of these prompts contains at least two objects and multiple or mixed types of attributes related to each object. We use the prompts generated in T2I-Benchmark~\citep{Huang2023T2ICompBenchAC} to make our results comparable. Stable Diffusion v1-4 (SDv1) and Stable Diffusion v2 (SDv2) are the T2I models adapted from Latent Diffusion Model (LDM)\cite{Rombach2021HighResolutionIS} and trained with large-scale data. Stable Diffusion XL (SDxl) is the most recent work that expands the original LDM with more parameters and introduces a refinement model to improve the image2image generation quality. SDxl also proposes novel conditioning schemes, including an extra text encoder, for better conditioning capability. All of our experiments are completed on an RTX3090 GPU server. 

\paragraph{Human Scoring Metrics}
We enlisted a group of annotators to score a total of 12,000 generated images over 1000 prompts. The scoring criteria were developed based on the severity of the hallucination phenomena observed in the images. We believe that for T2I tasks, understanding the entities mentioned in the text and their interrelations is fundamental. Therefore, the most serious errors in generated images are the omission of entities mentioned in the prompts or incorrect representation of the relationships between entities. Correspondingly, we consider inaccuracies in object attributes or the appearance of `extraneous objects' to be more tolerable errors in T2I tasks, as these errors appeared more frequent. 
Based on the aforementioned criteria, we established the following scoring system:
\begin{itemize}
    \item Generated images off-topic or unfathomable (1pt);
    \item Generated images with more than two missing objects or errors in the relationships between objects contrary to the text prompts (2pt);
    \item Generated images with two or fewer missing objects or errors in the relationships between objects contrary to the text prompts (3pt);
    \item Generated images with more than two object attribute errors not matching the text prompts (4pt)
    \item Generated images with two or less object attribute errors not matching the text prompts (5pt)
    \item Generated images containing objects not mentioned in the text (6pt)
    \item Generated images that perfectly align with the textual prompts (7pt)
\end{itemize}
It is important to note that in our evaluation, the presence of multiple objects of the same type in the generated images is not considered an `extraneous object' error. Moreover, if at least one object of the same type satisfies the attributes or relationships specified in the text, it is deemed correct. 

\subsection{Quantitative Analysis}
To demonstrate our method's comprehensive and multi-dimensional evaluation capabilities for composite T2I tasks, we present a comparison in Table~\ref{table:table1} between our approach and other existing evaluation metrics on images generated by different T2I models. Table~\ref{tab:table2} showcases our ability to accurately identify various types of hallucination errors as a composite evaluation method. These types encompass a diverse range of categories, from attribute errors to object missing.

\subsubsection{Comparison across evaluation metrics.} 
We employed CLIP-score~\cite{Hessel2021CLIPScoreAR,Radford2021LearningTV}, BLIP-score~\cite{Li2022BLIPBL}, LLM-score~\cite{Lu2023LLMScoreUT}, T2I-CompBench~\cite{Huang2023T2ICompBenchAC}, and our method to evaluate the total of 12,000 images generated from three different type of Stable diffusion models. We also include the average human evaluated scores for all of the images in the table as gold standard. The CLIP-score method encodes both the generated images and the prompt text with two independent encoders and calculates their cosine similarities as the final score. The BLIP model computes an ITM (Image-Text Matching) score to indicate the degree of correspondence between an image and text. This score is one of the objective functions during the model's pre-training phase. LLM-score first generates the caption of images and then compares their similarities using LLM As demonstrated in Table~\ref{table:table1}, we reported the average value for the normalized score from various evaluation methods. Although all scores have been normalized to the same range, we still observe distinct distributions among the values provided by different evaluation metrics. For example, the absolute scores from the T2ICompBench metric are significantly lower than those from other metrics. Therefore, to facilitate a more intuitive comparison of different metrics, we include the average Kendall and Spearman correlation coefficients between each score and the manual ratings in the last two columns of the table. Additionally, the manual scoring in the table's final row can serve as a reference for assessing the effectiveness of different evaluation metrics. 

From Tabel~\ref{table:table1}, we can spot several insights: 1. According to the human evaluation, SDxl~\cite{podell2023sdxl} receive the highest average score, compared to the other version of Stable Diffusion Models (SDM). This is reasonable as SDxl has more parameters and introduce a second text encoder. These results should not be viewed as a definitive comparison of the capabilities of different generative models, but rather as an indication of their ability to understand and reconstruct text. Moreover, the pattern observed in the manual scoring can be used to compare various evaluation metrics. Table~\ref{table:table1} shows that other models failed to demonstrate this pattern except for our method. As expected, the scores given by our method(with GPT-3.5) also show the highest correlation with human ratings. The aforementioned findings corroborate that our method exhibits a stronger correlation with human ratings compared to other methods.

\subsubsection{Categorizing Hallucination Types}
As previously mentioned, existing T2I evaluation methods are not adept at categorizing types of errors in generated images that do not align with the text. For methods like the CLIP-score, which rely solely on similarity as a metric, the result is a vague score, leaving it unclear exactly what types of errors are present in the generated images. We believe that for an evaluation metric, accurately pinpointing where the assessed model errs is key to guiding its improvement. Therefore, our method effectively categorizes different types of hallucination errors and quantifies them. As shown in Table~\ref{tab:table2}, our approach can distinguish four types of hallucination issues: attribute errors, relational errors, object omissions, and extraneous objects. Additionally, VQA models, such as T2ICompbench, can categorize hallucination issues by running separate modules. However, T2ICompbench does not perform as well as our method in classifying errors, and it is limited in the types of errors it can identify, failing to provide detailed, granular reasons for the errors. As demonstrated in Table~\ref{tab:table2}, our method achieves higher F1 scores across different hallucination types, surpassing the T2ICompbenchmark.  We report three variants of our method which adopt different LLMs (Gemini-1.5, GPT-3-turbo, and GPT-4) to be our question-answering base model. Nonetheless, due to the expensive API cost for GPT-4, we only applied the GPT-3-turbo and Gemini versions on full data, while we only recorded the GPT-4 variant's performance on about ten percent of our data that was randomly sampled from the 12,000 images. 
Our approach is particularly accurate in detecting attribute errors, missing object errors, and extraneous object errors, but its performance in identifying relational errors is more moderate. We analyze that this may be due to the multifaceted nature of expressing relationships, such as spatial positioning or verb forms. Specifically, verb-type relationships can lead to ambiguous interpretations in images. For instance, an image depicting a dog chasing a cat could align with various verb-relation in textual prompts like `chasing,' `playing,' or `running.' This subtle difference in expression makes accurately identifying the corresponding relationship challenging, and it is an issue we aim to address in our future work. 

\begin{table}[ht] 
    \centering
    \renewcommand{\arraystretch}{1.2}
    \resizebox{\columnwidth}{!}{%
    \begin{tabular}{c|cccc}
        \hline
        \textbf{Metrics}  & attribute & omission & relation & extraneous \\ \hline
        T2ICompbench     & 0.49  &    -    & 0.62  & -     \\ 
        \midrule
        Ours(w/Gemini)  & 0.45  & 0.79 & 0.61 & 0.36 \\
        Ours(w/GPT-3.5-turbo) & 0.52 & 0.76 & 0.62 & 0.33 \\
        Ours(w/GPT-4)          & \textbf{0.55}  & \textbf{0.81}  & \textbf{0.67} & \textbf{0.45}  \\ \hline

    \end{tabular}%
    }
    \caption{F1 score of different error types}
    \label{tab:table2}
\end{table}

\begin{figure*}[ht]
    \centering
    \begin{adjustbox}{center}
    \includegraphics[width=1.1\textwidth]{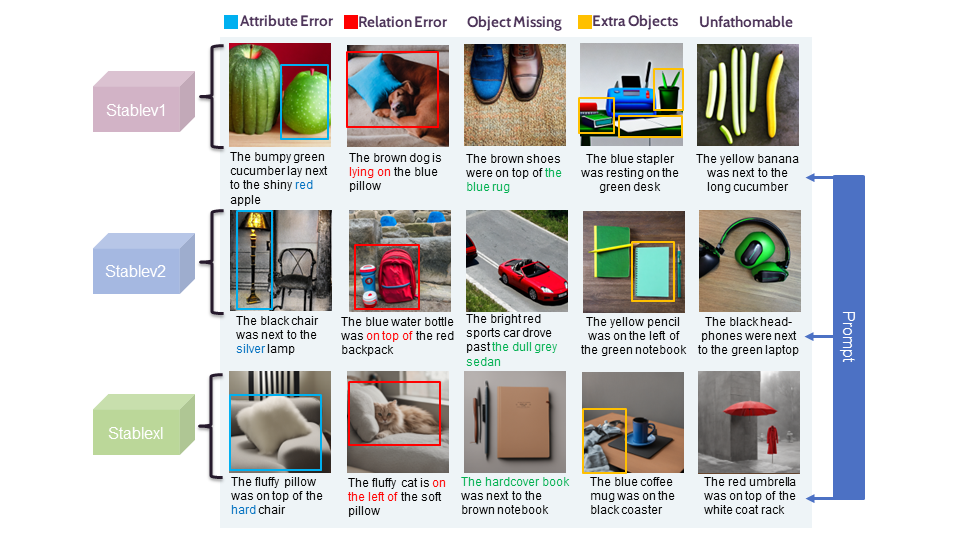}
    \end{adjustbox}
    \caption{Qualitative Evaluation: We select several hallucinated images not aligned with the prompts for each generative models. Each column indicate a specific type of hallucination issue. The last column are the images that totally off-topic to the prompts.}
    \label{fig:fig2}
\end{figure*}

\subsection{Ablation study}
To verify choosing graph-form knowledge representation is the best method for our task, we conduct an ablation study with an alternative knowledge representation method. We directly use caption sentences as knowledge base to do retrieval for question-answering and evaluate the generated contents. We select 1000 images from each model's output category and use the VLMs to generate captions from the synthesized images. Then, we provide the captions as context to the same LLM model we used before.
Moreover, we asked the exact same questions that applied to our method. Table~\ref{tab:ablation} shows that our method receives higher human correlation results than those using captions as a knowledge source. This result indicated that our method can retrieve the most relevant content from the graph triples. Compared to a complete knowledge graph, the captions sometimes omit some attributes, objects, and their relationships, even when we ask the models to provide captions in detail. This omission of crucial information will cause poor question-answering performance.

\begin{table}[ht] \large
\centering
\caption{Comparison between Caption-based and KG-based human score correlation}
\label{tab:ablation}
\resizebox{\columnwidth}{!}{%
\begin{tabular}{c|ccc}
\hline
\textbf{Methods} & Pearson & Kendall's $\tau$ & Spearman's $\rho$ \\[6pt] \hline
Caption-based(BLIP2-t5xxl )   & 0.18  & 0.15 &0.20  \\ [6pt]
Caption-based(Llava-13B)   & 0.16  & 0.13 &0.19  \\ [6pt]
Caption-based(Gemini)   & 0.21  & 0.17 &0.23  \\ [6pt]
\midrule
KG-based(GPT)   & \textbf{0.26}   & \textbf{0.23}  & \textbf{0.26} \\ [6pt]\hline
\end{tabular}%
}
\end{table}
\subsection{Qualitative Analysis}
As illustrated in Figure 2, we selected the most representative hallucinated examples categorized by the types of hallucinations for each model. From this figure, we can observe some insightful findings. First, the issue of attributes being bound to the wrong objects is prevalent across multiple models. This often stems from the ambiguity of references in the text. Clearly, current text encoders have not effectively resolved this issue, indicating a need for further exploration in this area. Another observation is that the generative model omits some objects from time to time. Based on our empirical understanding, the omission is caused by misunderstanding and confusion. For example, in Figure~\ref{fig:fig2} row two column three, the model probably confuse the `dully grey sedan' with the grey runway. 
Finally, in the last column, we exhibit several examples of generated images that our evaluators do not understand. For example, an image generated from the prompt: `a cucumber and a green banana' tends to blend the characteristics of bananas and cucumbers, leading to recognition failures for our detection model. 
These examples show that generating images from composite prompts is still a challenging task for current generative models. Furthermore, we hope our observations will inspire more work aimed at addressing the issues we have previously raised, thereby improving existing T2I models.


\section{Future works and Limiations}
As discussed earlier, our method struggles to effectively detect key objects in synthesized landscape images, possibly due to these being treated as background by GroundedSAM. We plan to address this issue in future work. Lastly, we believe that a more fine-grained and attribute-binding sensitive text encoder is urgently needed in current Text-to-Image (T2I) generative models, which will also be a direction for future research.

\end{document}